\documentclass[twoside,11pt]{article}
\usepackage{jair, rawfonts}

\usepackage{hyperref}

\hypersetup{
    colorlinks=true,
    linkcolor=black,
    filecolor=black,
    urlcolor=black,
    citecolor=black,
}

\usepackage{natbib}
\usepackage[most]{tcolorbox}
\usepackage{float}
\usepackage{booktabs, multirow}
\usepackage{subfig}
\usepackage{multicol}
\usepackage{url}

\newcommand*{\myfont}{\fontfamily{lmss}\selectfont}
\DeclareTextFontCommand{\textmyfont}{\myfont}

\begin{document}

\title{Fine-tuning Large Language Models for \\
       Adaptive Machine Translation}

\author{\name Yasmin Moslem \email {\scriptsize yasmin.moslem@adaptcentre.ie} \\
       \addr {\scriptsize ADAPT Centre, School of Computing, Dublin City University \\
       Dublin, Ireland}
       \AND
       \name Rejwanul Haque \email {\scriptsize rejwanul.haque@adaptcentre.ie} \\
       \addr {\scriptsize ADAPT Centre, Department of Computing, South East Technological University \\
       Carlow, Ireland}
       \AND
       \name Andy Way \email {\scriptsize andy.way@adaptcentre.ie} \\
       \addr {\scriptsize ADAPT Centre, School of Computing, Dublin City University \\
       Dublin, Ireland}
       }

\maketitle

\begin{abstract}
This paper presents the outcomes of fine-tuning Mistral 7B, a general-purpose large language model (LLM), for adaptive machine translation (MT). The fine-tuning process involves utilising a combination of zero-shot and one-shot translation prompts within the medical domain. The primary objective is to enhance real-time adaptive MT capabilities of Mistral 7B, enabling it to adapt translations to the required domain at inference time. The results, particularly for Spanish-to-English MT, showcase the efficacy of the fine-tuned model, demonstrating quality improvements in both zero-shot and one-shot translation scenarios, surpassing Mistral 7B's baseline performance. Notably, the fine-tuned Mistral outperforms ChatGPT ``gpt-3.5-turbo" in zero-shot translation while achieving comparable one-shot translation quality. Moreover, the zero-shot translation of the fine-tuned Mistral matches NLLB 3.3B's performance, and its one-shot translation quality surpasses that of NLLB 3.3B. These findings emphasise the significance of fine-tuning efficient LLMs like Mistral 7B to yield high-quality zero-shot translations comparable to task-oriented models like NLLB 3.3B. Additionally, the adaptive gains achieved in one-shot translation are comparable to those of commercial LLMs such as ChatGPT. Our experiments demonstrate that, with a relatively small dataset of 20,000 segments that incorporate a mix of zero-shot and one-shot prompts, fine-tuning significantly enhances Mistral's in-context learning ability, especially for real-time adaptive MT.

\end{abstract}

\section{Introduction}
\label{sec:intoduction}

Adaptive MT with LLMs is a task that involves employing the in-context learning feature of LLMs to adapt the MT output to approved similar translations \citep{Agrawal2023-SelectionMT,Moslem2023-AdaptiveMT} or terminology \citep{Ghazvininejad2023-DictionaryMT,Moslem2023-Terminology}. In this sense, feeding an LLM with extra context can help improve the translation quality and adherence to the domain terminology and style. By definition, in-context learning involves replicating text generation patterns without additional fine-tuning \citep{Brown2020-GPT-3}. Nevertheless, fine-tuning can still be an option to improve the in-context learning capability of LLMs.

The majority of current research that aims at enhancing MT through LLM training involves either: (i) pre-training an LLM to work with zero-shot MT \citep{Schioppa2023-LLM-MT-CrossLingual}; or (ii) fine-tuning LLMs with the main purpose of improving zero-shot capabilities \citep{Sia2022-LLM-MT,Alves2023-LLM-MT-Finetuning,Iyer2023-LLMs-Disambiguation,Jiao2023-Parrot,Xu2023-Llama-Fine-tuning,Zhang2023-MT-Efficient-Finetuning}. Moreover, there are several works that investigate pre-training or fine-tuning encoder-decoder MT models for adaptive MT \citep{Farajian2017-AdaptiveMT,Bulte2019-fuzzy,Xu2020-fuzzy}, and there is at least one work that compares this with using in-context learning of LLMs for adaptive MT \citep{Reinauer2023-MT-ICL}. However, there is still a need for research that instead investigates fine-tuning available open-source models to enhance their in-context learning ability for real-time adaptive MT and compares this to current approaches. To this end, these models can be fine-tuned to perform better at in-context learning scenarios, where special prompt templates incorporate in-domain sentences, phrases, or terminology. This direction can improve both translation quality and efficiency, especially as fewer examples might be required for in-context learning.

Among the benefits of fine-tuning open-source LLMs are efficient self-hosting. In other words, those who would like to serve their own LLMs for privacy reasons can utilise an open-source model, and efficiently achieve quality gains comparable to those of strong commercial LLMs. Moreover, so far, for very high-resource languages, LLMs can be used independently. However, for other languages, a hybrid approach using both conventional MT models and LLMs leads to better results, which means we have to deploy/use two models at translation time. If fine-tuning a small ``standalone'' LLM is possible for both regular (zero-shot) and adaptive (one-shot or few-shot) translation, this would be much more efficient. In addition, researchers can definitely build on this direction rather than having to rely on closed models. Open-source research can lead to more interpretability as we know better what is going on in the background.

In this work, we aim to investigate enhancing real-time adaptive MT with LLMs via fine-tuning on a mix of zero-shot and one-shot translation prompts. In the case of one-shot prompts, each new source segment is augmented with a similar translation pair (fuzzy match). Our experiments for the Spanish-to-English medical domain show that fine-tuning Mistral 7B on a small dataset, consisting of only 20,000 translation pairs, can improve it in-context learning capability for adaptive MT.

\section{Experimental Setup}
\label{sec:setup}

This section demonstrates our experiments with fine-tuning an LLM, namely Mistral 7B \citep{Jiang2023-Mistral}, for ``adaptive" MT. Hence, the model is not only fine-tuned for regular (zero-shot) translation, but also for adaptation to one fuzzy match (one-shot) at translation time. The experiments were conducted for Spanish-to-English medical adaptive MT. The code used for these experiments is publicly available.\footnote{\url{https://github.com/ymoslem/Adaptive-MT-LLM-Fine-tuning}}

\subsection{Data}
\label{sec:data}

\begin{figure}[H]
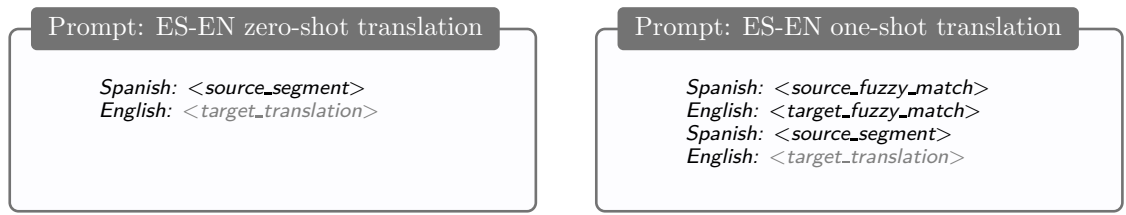

\captionsetup{font=footnotesize,labelfont=footnotesize}
\begin{multicols}{2}
\begin{tcolorbox}[enhanced,attach boxed title to top left={yshift=-3mm,yshifttext=-1mm,xshift=3mm},
  colback=blue!1!white,colframe=gray!90!black,colbacktitle=gray!90!black,boxrule=1pt,
  left=3pt, top=0pt,height=70pt, width=200pt,center,
  title=Prompt: ES-EN zero-shot translation,fonttitle=\small,
  boxed title style={size=small,colframe=gray!90!black} ]
  \begin{scriptsize}
  \textmyfont{\emph{
    \begin{itemize}
    \setlength\itemsep{-1.5ex}
    \item[] Spanish: $<$source\_segment$>$
    \item[] English: {\color{gray}$<$target\_translation$>$}
\end{itemize}
}}
\end{scriptsize}
\end{tcolorbox}

\begin{tcolorbox}[enhanced,attach boxed title to top left={yshift=-3mm,yshifttext=-1mm,xshift=3mm},
  colback=blue!1!white,colframe=gray!90!black,colbacktitle=gray!90!black,boxrule=1pt,
  left=3pt, top=0pt, height=70pt, width=200pt, center,
  title=Prompt: ES-EN one-shot translation,fonttitle=\small,
  boxed title style={size=small,colframe=gray!90!black} ]
  \begin{scriptsize}
  \textmyfont{\emph{
    \begin{itemize}
    \setlength\itemsep{-1.5ex}
    \item[] Spanish: $<$source\_fuzzy\_match$>$
    \item[] English: $<$target\_fuzzy\_match$>$
    \item[] Spanish: $<$source\_segment$>$
    \item[] English: {\color{gray}$<$target\_translation$>$}
\end{itemize}
}}
\end{scriptsize}
\end{tcolorbox}
\end{multicols}
\vspace{-1ex}
\caption{Zero-shot and one-shot prompts used for fine-tuning Mistral}
\label{fig:prompts}
\end{figure}

In this experiment, fine-tuning uses a mix of 10,000 segments with zero-shot prompts and 10,000 segments with one-shot prompts. The whole dataset was split into 19,000 segments for training the model and 1,000 randomly selected segments for validation while training. Fuzzy matches are extracted from a ``context dataset" including 50,000 translation pairs. The test dataset includes 10,000 sentences, and it has its own unique context dataset, which consists of 50,000 unique translation pairs. Figure \ref{fig:prompts} shows examples of zero-shot and one-shot prompts. The retrieval process of fuzzy matches is detailed in Section \ref{sec:information-retrieval}.

Originally, we mixed Spanish-to-English medical datasets from OPUS \citep{Tiedemann2012-OPUS}, namely ELRC \cite[et al.]{ELRC2019}, EMEA \citep{EMA2012}, SciELO \citep{Soares2018-SciELO}, and TICO-19 \citep{Anastasopoulos2020-TICO-19} datasets. Then we filtered\footnote{Scripts are available at: \url{https://github.com/ymoslem/MT-Preparation}} the resulted dataset, to exclude duplicates and too long segments.\footnote{As NLLB supports a maximum token length of 512 \textit{tokens}, we excluded any segment whose source or target text longer than 70 \textit{words} to also take into account the one-shot case that will augment another segment to the original one. As the context window of Mistral is much larger (8K tokens), it is theoretically possible to translate longer segments.}
The whole dataset includes 1,868,505 segments before filtering, and 922,343 segments after filtering.\footnote{We observe that almost two-thirds of the EMEA dataset are duplicates.} However, we used only part of it for this preliminary experiment. In the future, we would like to increase the size of the training data and compare the performance. Nevertheless, achieving these results (cf. Table \ref{tab:mistral-finetuning}) with such a small dataset (cf. Section \ref{sec:data}) shows how promising this approach is.

\subsection{Information Retrieval}
\label{sec:information-retrieval}

For indexing and retrieval of fuzzy matches, we use Sentence-Transformers \citep{Reimers2019-SentenceTransformers} and \textit{Faiss} \citep{Johnson2019-Faiss}, with a multilingual model to generate the embeddings for the datasets and later to extract fuzzy matches through semantic search.

\paragraph{Embedding:} To encode all the translation segments into embeddings, we employ a \textit{multilingual} model, namely Microsoft's \textit{``Multilingual-MiniLM-L12-H384"} \citep{Wang2020-MS-MiniLM}. These embeddings will be used later for both indexing and retrieval. The step of generating embeddings can be implemented with the Sentence-Transformers library.\footnote{\url{https://www.sbert.net/}}

\paragraph{Indexing:} For indexing, we use Faiss,\footnote{\url{https://github.com/facebookresearch/faiss}} a library for efficient similarity search and clustering of dense vectors. We train an \emph{IndexIVFFlat} index, which uses \emph{IndexFlatL2} as a quantiser.\footnote{\url{https://github.com/facebookresearch/faiss/wiki/Faster-search}}
The embedding size is 384, which is the same as the embedding size of model used.
For the number of clusters at indexing time, the \emph{nlist} parameter was set to 4096, while the number of clusters to explore at search time \textit{nprobe} was set to search for nearest neighbours in 32 clusters.\footnote{According to Faiss' guidelines for choosing an index, the number of clusters is recommended to be between \emph{4*sqrt(N)} to \emph{16*sqrt(N)}, where \emph{N} is the size of the dataset. As the ``context dataset" includes 50,000 segments, we might experiment in the future with increasing the number of clusters  while training an index. Obviously, this would depend on the available computational resources.}

\paragraph{Semantic Search:} This step computes the cosine similarity between the query and all the documents in the corpus based on their embeddings, and retrieves the top $k$ matching entries. In this case, our query is each source segment, and the corpus is the unique ``context dataset" (cf. Section \ref{sec:data}) leveraged to extract fuzzy matches.

\paragraph{Re-ranking:} The embedding step is usually done by a more efficient model to retrieve the most probable results (e.g. the top 100 results). Later, a re-ranker based on a cross-encoder is used to score the relevancy of all candidates for the given search query. This dual process is supposed to improve both efficiency and quality of the retrieved documents. However, we did not apply this step in our experiments.

\subsection{Fine-tuning}
\label{sec:finetuning}

We used QLoRA \citep{Hu2021-LoRA,Dettmers2023-QLoRA} for efficient fine-tuning with 4bit quantization, with Hugging Face Transformers.\footnote{\url{https://github.com/huggingface/transformers}} Fine-tuning was for one epoch, which revealed better results than fine-tuning for 4 epochs. Quantization configuration through \textit{BitsAndBytes} includes: {\small \textit{load\_in\_4bit=True}, \textit{bnb\_4bit\_quant\_type``nf4"}, \textit{bnb\_4bit\_use\_double\_quant=True}, and \textit{bnb\_4bit\_compute\_dtype=torch.bfloat16}}. LoRA configuration was set via the PEFT library\footnote{\url{https://github.com/huggingface/peft}} as follows: the dimension of the low-rank matrices \textit{r=64}, the scaling factor for the weight matrices \textit{lora\_alpha=16}, dropout probability of the LoRA layers \textit{lora\_dropout=0.1}, and without training the bias parameters for better performance \textit{bias=``none"}.
%\footnote{\url{https://huggingface.co/docs/peft/task\_guides/token-classification-lora}}
Training arguments include: batch size for training and evaluation 32 examples, {\small \textit{warmup\_steps=0.03}, \textit{learning\_rate=2e-3}, \textit{lr\_scheduler\_type=``constant"}, and \textit{bf16=True}}.
Both training and inference utilise Google Colab Pro+ with one GPU \textit{NVIDIA A100-SXM4-40GB}.

\subsection{Inference}

For inference, we experimented with a number of models including NLLB-200 \citep{NLLB2022} whose architecture is encoder-decoder Transformer \citep{Vaswani2017-attention} as well as ChatGPT \cite{Brown2020-GPT-3,Ouyang2022-InstructGPT} and Mistral 7B, which are autoregressive decoder-only Transformer-based LLMs. Mistral 7B was used both without fine-tuning and after fine-tuning on a mix of zero-shot and one-shot translation prompts.

\paragraph{Mistral 7B:} For inference (translation), we converted both the baseline and our fine-tuned models of Mistral to the CTranslate2 \footnote{\url{https://github.com/OpenNMT/CTranslate2}} \citep{Klein2020-Efficient} format (with 8int quantization) for more efficiency. We employed greedy search by setting \textit{sampling\_topk=1} and added the new line \verb|\n| character to \textit{end\_token} to avoid overgeneration. Mistral through CTranslate2 translates the zero-shot test dataset that includes 10,000 sentences in 2–3 minutes (approx. 80 segments/second). The time almost doubles for the one-shot test dataset. Figure \ref{fig:prompts} illustrates the prompts used for both inference and fine-tuning. Two versions of Mistral were tested, the baseline model without fine-tuning, and the model we fine-tuned on a mix of zero-shot and one-shot translation prompts. Table \ref{tab:mistral-finetuning} shows translation evaluation results of both models. Fine-tuning Mistral on a mix of zero-shot and one-shot prompts improved both its regular translation quality (i.e. when only the source text is available) and adaptive translation quality (in this case, when one fuzzy match is provided) at inference time.

\paragraph{ChatGPT:} The model used is ``gpt-3.5-turbo'' with \textit{temperature=0.3} and \textit{top\_p=1}. Requests are sent in batches of 20 segments, and the \textit{max\_tokens} argument was set as the largest number of words per source segment in a batch multiplied by 4, which is a rough number that can be increased or decreased based on the language and model. When the source text is augmented with one fuzzy match, the translation quality is improved by several points across all the automatic evaluation metrics. These quality gains are in line with results from previous work \citep{Moslem2023-AdaptiveMT} although in this work we are using a different dataset. However, it is worth noting that although batch processing was employed, there is no guarantee of the generation time with ChatGPT, which can range from several minutes to a couple of hours. In this sense, we observe that Mistral 7B is much more efficient, especially when used with CTranslate2.

\paragraph{NLLB-200:} In this set of experiments, the NLLB model was used as is, i.e. without fine-tuning. The first two rows of Table \ref{tab:mistral-finetuning} show the evaluation scores of using NLLB 3.3B for translation without a fuzzy match (zero-shot) and with a fuzzy match (one-shot). The same test dataset and its unique context dataset were used; however, fuzzy matching augmentation was done differently to match the architecture of the NLLB model. Each source sentence was augmented with its best fuzzy match, and the two sentences were separated by the language code of the source language (in this case ``spa\_Latn"). As NLLB was mainly pre-trained on sentences, we had to add an extra token that usually comes at the beginning of sentences, such as a bullet point (``{\tiny •}") after the language code between the two source sentences. Hence, the target fuzzy match was fed to the model as a prefix augmented by the target language code (in this case ``eng\_Latn") and the extra token. In this sense, the model was encouraged to complete the translation through teacher-forcing \citep{Williams1989-TeacherForcing}, i.e. using the ground truth as input, instead of the model output. In other words, the model is not required to translate the target fuzzy match, but rather to use the provided translation as is to guide the translation of the new untranslated source sentence. Translation arguments include: \textit{batch\_type=``tokens"}, \textit{max\_batch\_size=2024}, \textit{beam\_size=2}, \textit{min\_decoding\_length=2}, and \textit{max\_decoding\_length=512}. Although there is a marginal improvement given the BLEU score \citep{Papineni2002-BLEU}, the performance degrades according to the other reported automatic evaluation metrics, chrF++ \citep{Popovic2017-chrF++}, TER \citep{Snover2006-TER} and COMET \citep{Rei2020-COMET}. The fact that NLLB was trained to translate individual sentences rather than a series of sentences or full documents could be the main reason for this. In the future, we would like to experiment with fine-tuning NLLB with fuzzy matching augmentation and compare the results.

\newpage
\section{Results}
\label{results}

This section demonstrates the results of fine-tuning Mistral 7B for ``adaptive" MT. As detailed in Section \ref{sec:setup}, the model is not only fine-tuned on a mix of zero-shot and one-shot translation prompts to improve enhance real-time adaptive MT. In other words, it was fine-tuned to boost the LLM ability to adapt it output to the required domain at translation time. The experiment was conducted for Spanish-to-English medical adaptive MT.

\begin{table}[htp]
\setlength{\tabcolsep}{4pt}
    \captionsetup{font=footnotesize,labelfont=footnotesize}
    \centering
    \begin{small}
    \begin{tabular}{ccccccc}
    \toprule
    \textbf{Lang} & \textbf{Model} & \textbf{Context} & \textbf{BLEU ↑} & \textbf{chrF++ ↑} & \textbf{TER ↓} & \textbf{COMET ↑} \\ \midrule 
    \multirow{8}{*}{\textbf{ES-EN}} & \multirow{2}{*}{NLLB 3.3B} & Source only (zero-shot) & \underline{47.02} & 68.82 & 43.43 & 66.46 \\
     &  & + Fuzzy (one-shot) & 47.42 & 68.77 & 45.26 & 64.57 \\ \cmidrule{2-7}
     & ChatGPT & Source only (zero-shot) & 44.65 & 68.36 & 44.28 & 74.48 \\
     & {``\footnotesize gpt-3.5-turbo"} & + Fuzzy (one-shot) & 48.34 & 70.54 & 40.80 & \textbf{80.25}* \\ \cmidrule{2-7}
     & \multirow{2}{*}{Mistral 7B} & Source only (zero-shot) & 42.88 & 66.03 & 46.54 & 69.56 \\
     &  & + Fuzzy (one-shot) & 47.35 & 69.25 & 42.53 & 76.37 \\ \cmidrule{2-7}
     & Mistral 7B & Source only (zero-shot) & 46.71 & \underline{69.55} & \underline{41.81} & \underline{77.44} \\
     & ``\textbf{Fine-tuned}" & + Fuzzy (one-shot) & \textbf{49.69} & \textbf{70.89} & \textbf{40.08} & \textbf{79.62} \\ \bottomrule
    \end{tabular}
    \end{small}
    \caption{Comparing adaptive MT with NLLB-200 3.3B, ChatGPT, and Mistral 7B (before and after fine-tuning). Our fine-tuned version of Mistral demonstrates translation quality gains that are on par with the task-oriented NLLB model for zero-shot translation, and outperform both NLLB and ChatGPT for one-shot adaptive MT with one fuzzy match.}
    \label{tab:mistral-finetuning}
\end{table}

In Table \ref{tab:mistral-finetuning}, the last two rows show the results for fine-tuning Mistral 7B. The rest of the results are for baselines, i.e. without fine-tuning. As illustrated, fine-tuning has led to quality improvements in terms of both zero-shot and one-shot translation. The fine-tuned version of Mistral outperforms its own baseline (i.e. without fine-tuning) for both zero-shot and one-shot translation. Zero-shot translation of the fine-tuned Mistral outperforms ChatGPT ``gpt-3.5-turbo", while one-shot translation quality of fine-tuned Mistral is on par with that of ChatGPT. Zero-shot translation of the fine-tuned Mistral is on par with NLLB 3.3B, while one-shot translation quality of the fine-tuned Mistral outperforms that of NLLB 3.3B. To conclude, fine-tuning an efficient LLM like Mistral 7B helps to produce a high-quality zero-shot translation comparable to that of MT task-oriented models such as NLLB 3.3B, while achieving adaptive gains of one-shot translation on par with commercial LLMs such as ChatGPT ``gpt-3.5-turbo".

\section{Conclusion and Future Work}

In this work, we showed how fine-tuning a general purpose LLM such as Mistral 7B can improve its in-context learning ability, especially for real-time adaptive MT. Moreover, such translation quality gains were achievable through fine-tuning using a relatively small dataset (20,000 segments). Incorporating a mix of zero-shot and one-shot prompts in the training data helps improve both regular zero-shot translation, and one-shot translation that incorporates a fuzzy match. It is worth noting that Mistral 7B is much more efficient than ChatGPT, which is an added benefit in production scenarios.

In the future, we would like to experiment with other domains and language pairs, including low-resource languages. As currently there are several multilingual LLMs such as BLOOM (46 languages) \citep{BLOOM2022}, Falcon (EN-DE-ES-FR) \citep{Penedo2023-Falcon}, larger versions of Mistral/Mixtral (EN-DE-ES-FR-IT) \citep{Jiang2023-Mistral}, Jais (AR-EN) \citep{Sengupta2023-Jais}, Baichuan (ZH) \citep{Yang2023-Baichuan}, and Qwen (ZH) \citep{Bai2023-Qwen}, it can be insightful to apply the same approach with these models. Furthermore, as we experimented with NLLB-200 without fine-tuning, we would like to experiment with fine-tuning for fair comparison. While we fine-tuned Mistral on a small dataset, it is recommended to experiment with fine-tuning on more data, especially from the same domain. It can also be helpful to incorporate different types of prompts, such as few-shot prompts, and terminology-based prompts.

\acks{This work is supported by the Science Foundation Ireland (SFI) Centre for Research Training in Digitally-Enhanced Reality (d-real) under Grant No. 18/CRT/6224, the ADAPT Centre for Digital Content Technology under SFI's Grant No. \mbox{13/RC/2106\_P2}, and \mbox{Microsoft} Research.
}

\vskip 0.2in
\bibliography{paperpile}

\end{document}